\documentclass[10pt,twocolumn,letterpaper]{article}

\usepackage[letterpaper,left=0.8125in,right=0.8125in,top=1in,bottom=1in,columnsep=0.3125in]{geometry}

\usepackage[utf8]{inputenc}
\usepackage[T1]{fontenc}
\usepackage{mathptmx}
\usepackage{microtype}

\usepackage{amsmath,amssymb,amsthm}
\usepackage{bm}
\usepackage{booktabs}
\usepackage{multirow}
\usepackage{arydshln}
\usepackage{graphicx}
\usepackage{subcaption}
\usepackage{xcolor}
\usepackage{pgfplots}
\pgfplotsset{compat=1.18}

\newtheorem{proposition}{Proposition}

\usepackage[numbers,sort&compress]{natbib}

\definecolor{linkblue}{rgb}{0.21,0.49,0.74}
\usepackage[breaklinks,colorlinks,allcolors=linkblue]{hyperref}
\usepackage[capitalize]{cleveref}
\crefname{section}{Sec.}{Secs.}
\Crefname{section}{Section}{Sections}
\crefname{table}{Tab.}{Tabs.}
\Crefname{table}{Table}{Tables}
\crefname{figure}{Fig.}{Figs.}
\Crefname{figure}{Figure}{Figures}
\crefname{equation}{Eq.}{Eqs.}
\Crefname{equation}{Equation}{Equations}

\title{Robustness Verification against Adversarial Patch Attacks \\via Group-Lasso Semidefinite Relaxation}

\newcommand{\authorbox}[4]{%
  \begin{tabular}[t]{@{}c@{}}
    {\large #1}\\
    \small #2\\
    \small #3\\
    {\tt\small #4}
  \end{tabular}}

\author{%
  \begin{tabular}[t]{@{}c@{\hspace{1.5em}}c@{}}
    \authorbox{Yukina Nakagawa}{Institute of Science Tokyo}{2-12-1 Ookayama, Meguro-ku, Tokyo 152-8550, Japan}{nakagawa.y.367d@m.isct.ac.jp}
    &
    \authorbox{Takami Sato}{Keio University}{2-15-45 Mita, Minato-ku, Tokyo 108-8345, Japan}{takami.sato@yos.elec.keio.ac.jp}
    \\[1.5em]
    \authorbox{Ken Kobayashi}{Institute of Science Tokyo}{2-12-1 Ookayama, Meguro-ku, Tokyo 152-8550, Japan}{kobayashi.k@iee.eng.isct.ac.jp}
    &
    \authorbox{Kazuhide Nakata}{Institute of Science Tokyo}{2-12-1 Ookayama, Meguro-ku, Tokyo 152-8550, Japan}{nakata.k@iee.eng.isct.ac.jp}
  \end{tabular}
}
\date{}

\begin{document}
\newlength{\subfigImgHeight}
\setlength{\subfigImgHeight}{0.19\columnwidth}
\maketitle

\begin{abstract}
Robustness verification is essential for ensuring the safety and security of neural networks, but current methods focus heavily on $L_p$ norm-ball attacks, which do not adequately model typical real-world threats that involve physical perturbations, such as adversarial patch attacks. A naive approach requires computationally prohibitive enumeration of all possible patch positions. To address this challenge, we propose a novel neural network verification method specialized for patch-attack threats. Instead of enumerating patch locations, we first model the patch attack with group-wise sparsity and formulate verification as a mixed-integer semidefinite program (MISDP) by applying semidefinite program (SDP) relaxation. To further relax the problem to a continuous optimization problem, we propose three relaxations, including Group-Lasso relaxation, that remove the integer variables while preserving the sparse, group-wise structure of the original attack, enabling substantially more efficient verification. Our empirical analysis demonstrates that the proposed relaxations require less than 3\% of the computation time of the original MISDP, while Group-Lasso relaxation incurs only a 0.34\% average increase in the objective value over the original formulation. Furthermore, Group-Lasso achieves a verification rate of 91\%, compared with $\leq$70\% for prior methods.
\end{abstract}

\section{Introduction}
\label{sec:intro}

\begin{figure}[t]
  \centering
  \includegraphics[width=\linewidth]{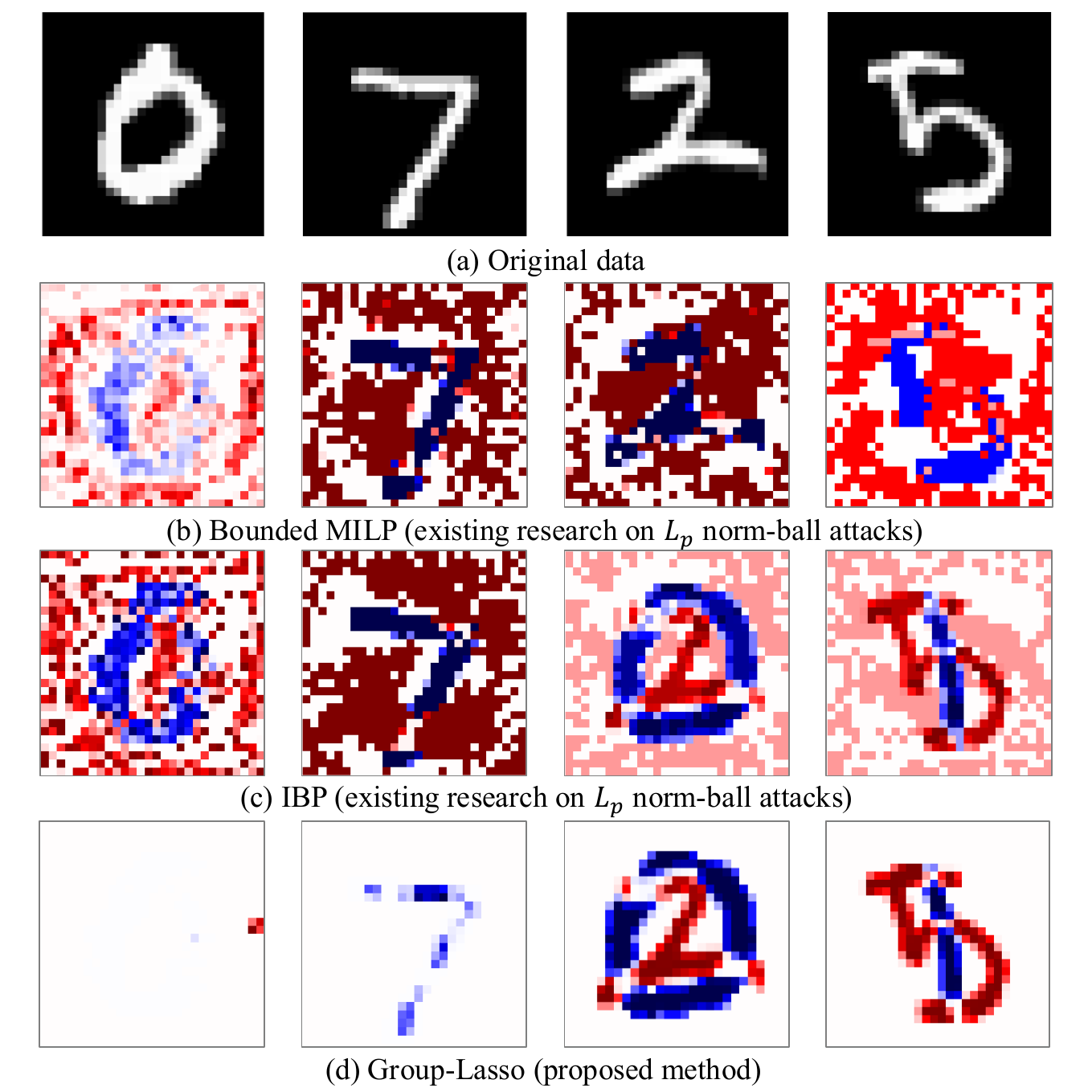}
  \caption{Adversarial perturbations generated during verification. Red and blue mean positive- and negative-direction perturbations, respectively. Compared to existing baselines (MILP and IBP) that spread perturbations across the entire image, our proposed Group-Lasso relaxation restricts perturbations to localized foreground regions, closely matching the localized patch attack threat model.}
  \label{fig:milp_ibp_glasso}
\end{figure}

Deep Neural Networks (DNNs) have achieved remarkable success in many fields and have been widely deployed in safety-critical vision applications such as autonomous driving~\cite{bojarski2016end,pomerleau1989alvinn} and medical image analysis~\cite{esteva2017dermatologist,wu1993artificial}. However, DNNs are widely known to be vulnerable to adversarial attacks~\cite{szegedy2014intriguing,goodfellow2015explaining,carlini2017towards} that intentionally modify an input with subtle malicious perturbations to induce an incorrect prediction. The vulnerability to adversarial attacks has raised critical concerns about their reliability and security, spurring extensive research into DNN verification~\cite{tjeng2019evaluating,wong2018provable,ehlers2017formal,katz2017reluplex}.

DNN verification has traditionally focused on $L_p$ norm-ball attacks. For an input image $\bm{x}^\ast$, the verifier considers all possible perturbed images $\bm{x}$ under a perturbation radius $\epsilon$, i.e., $||\bm{x}-\bm{x}^\ast||_p\leq\epsilon$, and determines whether the model prediction remains unchanged.  This verification problem is generally solved as an optimization problem. While the $L_p$ norm-ball assumption can cover broader types of perturbations, it does not adequately represent the threat model of practical physical-world attacks~\cite{eykholt2018physical,brown2017adversarial} in which an attacker modifies only a localized region, for example by attaching a sticker or placing a patterned object on part of an image. A naive approach to patch attacks enumerates possible patch locations and solves a verification problem for each candidate location, but the computational cost grows exponentially with the number of possible attack locations. 

To address this limitation, we propose a novel DNN verification method specialized for patch attacks. Instead of enumerating patch locations, we model the patch attack with group-wise sparsity~\cite{fan2020sparse}. We then incorporate the patch position as a variable within a single mixed-integer program (MIP), which is an exact model but is not practical due to its high combinatorial complexity.
We thus relax the problem with an SDP relaxation~\cite{lan2023semidefinite} and three proposed relaxations for the group-wise patch attack.
We propose a full-group naive relaxation that permits perturbations in every group, an $L_2$-norm relaxation that replaces the group-wise constraints with a global $L_2$ constraint, and a Group-Lasso relaxation that constrains the sum of the $L_2$ norms of the group-wise perturbations. All three relaxed problems provide a valid guarantee for the original attack model.

We evaluate the proposed formulations from both computational and verification perspectives. We found that all three relaxations require less than 3\% of the computation time of the original problem in the evaluated settings. Group-Lasso relaxation achieves an average objective increase of only 0.34\% in the evaluated setting, substantially outperforming the Full-Group and $L_2$-norm relaxations. These results indicate that preserving the group structure of the attack is important for obtaining a tight relaxation. Group-Lasso relaxation also demonstrated high verification capability; while existing methods achieve verification rates of $\leq$70\%, it was able to verify 91\% of the samples on the adversarially trained model.
We conduct qualitative analysis and confirm that the perturbations considered in Group-Lasso relaxation are locally grouped and concentrated around the digit character, as shown in \cref{fig:milp_ibp_glasso}.

Our main contributions are summarized as follows:
\begin{itemize}
\item We are the first to formulate DNN verification specialized for patch attacks with group-wise sparsity modeling. We first formulate it as an exact MIP problem and design an effective relaxation enabling efficient verification.

\item We design three relaxations for the group-wise modeling with theoretical soundness guarantees. In particular, Group-Lasso relaxation explicitly preserves the group-sparse structure and provides a substantially tighter approximation than the simpler relaxations.

\item We demonstrate the high efficiency and effectiveness of our proposed methods, which require at most 3\% of the computation time of the original MISDP. Group-Lasso relaxation differs from the exact formulation by only 0.34\% in objective value and achieves a verification rate of up to 91\%, outperforming existing methods, whose verification rates are at most 70\%.
\end{itemize}

\section{Related Work}
\label{sec:background}

\subsection{Adversarial Attacks}
\label{subsec:patch_defenses}

Deep Neural Networks (DNNs) are widely known to be vulnerable to adversarial attacks~\cite{szegedy2014intriguing,goodfellow2015explaining,carlini2017towards}, where carefully crafted perturbations alter model predictions while remaining imperceptible to humans. To bound perturbation magnitude, traditional attacks are typically formulated within an $L_p$-norm ball, such as $L_\infty$~\cite{madry2018towards} or $L_2$~\cite{carlini2017towards} balls.

Beyond global perturbation, recent work has demonstrated that effective attacks can be realized by perturbing only a localized region of the input image~\cite{liu2019dpatch,eykholt2018physical,sharif2016accessorize}. We refer to this threat model as \emph{patch attacks}.
Attacks based on $L_p$ norm balls perturb the entire image, which would require physically perturbing everything within the camera's field of view, an unrealistic requirement in the physical world. An attacker would therefore need to compromise the system itself. In contrast, patch attacks allow an adversary to introduce localized physical perturbations by placing patches within the camera's field of view~\cite{brown2017adversarial,sato2021dirty} or attaching stickers to target objects~\cite{eykholt2018physical}, without compromising the system. Owing to this practicality in real-world scenarios, patch attacks have drawn significant attention.

This emerging threat model has motivated the development of certified defenses against patch attacks~\cite{chiang2020certified, xiang2022patchcleanser, xiang2021patchguard, levine2020derandomized, metzen2021efficient}. Chiang et al.~\cite{chiang2020certified} proposed a certified defense that enumerates possible patch locations and performs verification for each location. Levine et al.~\cite{levine2020derandomized} developed a deterministic certification approach that restricts the set of possible patch regions and verifies the resulting cases. BagCert~\cite{metzen2021efficient} avoids explicit enumeration of all patch locations by partitioning the input image into multiple regions and performing certified defense for CNNs. However, existing approaches focus on developing certified defenses, not verification for all possible attacks. None of the prior studies has studied DNN verification methods specialized for patch attacks. This gap motivates our study.

\subsection{DNN Verification Problem}
\label{subsec:dnn_verification}

DNN verification aims to determine whether a model preserves its prediction for all inputs within a specified perturbation set. Given an input image $\bm{x}^*$ with correct class label $y^*$, the model $f:\mathbb{R}^n \to \mathbb{R}^C$ is certified as robust against the attack of interest if its predicted class remains unchanged throughout the perturbation set $\mathcal{X}$ centered at $\bm{x}^*$. We formulate the DNN verification problem as the following decision problem:

\begin{align}
    \label{eq:verification_abst}
    \forall \bm{x}\in \mathcal{X}, \;\operatorname*{arg\,max}_{i\in \{1,2,...,C\}} f(\bm{x})=\operatorname*{arg\,max}_{i\in \{1,2,...,C\}} f(\bm{x}^*)=y^*.
\end{align}

For an $L_p$ norm ball, the perturbation set can be defined as follows: 
\begin{align}
\mathcal{X}_{L_p}
=
\left \{
\bm{x}\mid
||\bm{x}-\bm{x}^\ast||_p\leq\epsilon
\right \}.
\label{eq:lp_ball}
\end{align}

DNN verification has been studied through both complete and incomplete verification approaches. Complete verification~\cite{ehlers2017formal,lomuscio2017approach,botoeva2020efficient,katz2017reluplex,tjeng2019evaluating,bastani2016measuring} seeks to solve Problem~\cref{eq:verification_abst} exactly, typically using mixed-integer programming (MIP)~\cite{tjeng2019evaluating,wong2018provable} or satisfiability modulo theories (SMT)~\cite{ehlers2017formal,katz2017reluplex}. Although these methods can provide exact verification, their computational cost is generally unaffordable for practical use.

Incomplete verification instead replaces the original verification problem with a relaxed problem that is computationally more tractable. The relaxed problem is constructed as a necessary condition for the original verification problem: consequently, if the relaxed problem is verified, then the original DNN verification problem is also verified. However, this relaxation can introduce false negatives, meaning that a model may be certifiably robust under complete verification even when the incomplete verifier fails to establish robustness.

\section{Methodology}
\label{sec:methodology}

We first formalize the patch-attack threat model using group-wise sparsity (\cref{subsec:threat_model}) and derive an exact mixed-integer program that jointly optimizes the perturbation and patch location (\cref{subsec:patch_attack_complete_veri}). We then apply an SDP relaxation to obtain a more tractable MISDP (\cref{subsec:incomplete_sdp_relax}), and finally propose three relaxations that remove the remaining combinatorial structure (\cref{sec:relaxations}).

\subsection{Patch Attack Modeling}
\label{subsec:threat_model}

We model the patch attack with group-wise sparsity~\cite{fan2020sparse}. 
Let the pixel index set of an input image be $\mathcal{P} = \{1, 2, \ldots, n_0\}$.
We partition $\mathcal{P}$ into $m$ non-overlapping groups:
\begin{subequations}
\label{eq:subregion_partition}
\begin{align}
    &\mathcal{G} = \{1, 2, \ldots, m\}, \label{eq:group_index}\\
    &\mathcal{P}_i \cap \mathcal{P}_j = \emptyset \quad (i, j \in \mathcal{G},\; i \neq j), \label{eq:group_disjoint}\\
    &\bigcup_{g \in \mathcal{G}} \mathcal{P}_g = \mathcal{P}. \label{eq:group_cover}
\end{align}
\end{subequations}
where $\mathcal{P}_g$ denotes the set of pixels belonging to the group $g$. \cref{eq:group_disjoint} and \cref{eq:group_cover} respectively guarantee the non-overlap and the coverage of the partition.

The attacker can select at most $k$ groups to perturb. 
The verifier cannot know which groups are selected. The perturbation applied to each selected group is bounded by $\epsilon$. Let $\bm{G}\in \{0,1\}^{m}$ denote the group-selection vector, where $G^{(g)}=1$ indicates that group $g$ is attacked. For an original input $\bm{x}_0^\ast$, the corresponding perturbation set is defined as
\begin{subequations}
\label{eq:gwise_perturb_set}
\begin{align}
    \| \bm{x}_0^{(g)} - \bm{x}_0^{*(g)} \|_2 &\leq \epsilon G^{(g)}
        \quad  (g \in \mathcal{G}), \label{eq:pert_group} \\
    \mathbf{1}^\top \bm{G} &= k, \label{eq:pert_card} \\
    \bm{G} &\in \{0, 1\}^m. \label{eq:pert_bin}
\end{align}
\end{subequations}
The vectors $\bm{x}_0^{(g)}$ and $\bm{x}_0^{\ast(g)}$ denote the sub-vectors of the perturbed and original images restricted to the index set $\mathcal{P}_g$ of group $g$. The first constraint allows a perturbation of radius at most $\epsilon$ only for an attacked group, while the second constraint restricts the attacker to at most $k$ groups. The binary vector $\bm{G} \in \{0,1\}^m$ indicates which groups are attacked, where $G^{(g)} = 1$ if group $g$ is attacked and $G^{(g)} = 0$ otherwise.

\subsection{Patch Attack Complete Verification Modeling}
\label{subsec:patch_attack_complete_veri}

Given the perturbation set defined in \cref{eq:gwise_perturb_set}, robustness verification can be formulated as an optimization problem that jointly searches for the adversarial perturbation and its attack location. For a given target class, we maximize the difference between the target and true class scores:
\begin{subequations}
\label{eq:abstract_gwise_verification_problem}
    \begin{align}
    \max_{\{\bm{x}_i\}_{i=0}^{L}, \bm G}&\bm{x}_L[j]-\bm{x}_L[j^*]\\
    \text{s.t.}\quad&\bm{x}_L = f(\bm{x}_0), \\
    &\| \bm{x}_0^{(g)} - \bm{x}_0^{*(g)} \|_2 \leq \epsilon G^{(g)},
        \quad \forall g \in \mathcal{G}, \label{eq:mip_pert_group} \\
    &\mathbf{1}^\top \bm{G} = k, \label{eq:mip_pert_card} \\
    &\bm{G} \in \{0, 1\}^m, \label{eq:mip_pert_bin}
    \end{align}
\end{subequations}
where $j^*$ and $j~(j\neq j^*)$ denote the true and target labels, and $\bm{x}_L[j]$ represents the $j$-th entry of the vector $\bm x_L$. 
$\bm{x}_0$ denotes the perturbed input and $\bm{G}$ determines the attacked groups. 
If the optimal value is non-positive, the model is verified to be robust against the specified target class; otherwise, the existence of a patch attack cannot be ruled out.

For an $L$-layer ReLU network, \cref{eq:abstract_gwise_verification_problem} can be formulated as an extension of the existing mixed-integer program (MIP)-based complete verification~\cite{katz2017reluplex}:
\begin{subequations}
\label{eq:gwise_verification_problem}
\begin{align}
    \max_{\{\bm{x}_i\}_{i=0}^{L},\bm{G}}&\bm{c}^\top \bm{x}_L+c_0\\
    \mathrm{s.t.} \quad
        &\bm{x}_{i+1}\geq \bm 0,\quad(i \in \mathbb{I}_{[0,L-1]}),\label{subeq:relu_constraint_begin}\\
        &\bm{x}_{i+1}\geq W_i\bm{x}_i+\bm{b}_i\quad(i \in \mathbb{I}_{[0,L-1]}),\\
        &\bm{x}_{i+1}\leq (1-\bm{z}_i)M+W_i\bm{x}_i+\bm{b}_i~(i \in \mathbb{I}_{[0,L-1]}),\label{subeq:relu_constraint_2}\\
        &\bm{x}_{i+1}\leq M\bm{z}_{i}\quad (i \in \mathbb{I}_{[0,L-1]}),\label{subeq:relu_constraint_end}\\
        &\bm 0\leq \bm{x}_0\leq \bm 1,\label{subeq:pixel_bounds}\\
        &\bm{z}_i \in\{0,1\}^{n_i}\quad (i \in \mathbb{I}_{[0,L-1]}),\\
        & \| \bm{x}_0^{(g)} - \bm{x}_0^{*(g)} \|_2 \leq \epsilon G^{(g)}\quad (g \in \mathcal{G}),\label{subeq:input_perturbation_constraint_1} \\
        &\mathbf{1}^\top \bm{G} = k,
            \quad \bm{G} \in \{0,1\}^m\label{subeq:input_perturbation_constraint_2}.
\end{align}
\end{subequations}
where  $\bm{c}^\top = W_L(j,\ ;)-W_L(j^*,\ ;)$ and $c_0 = \bm{b}_L[j]-\bm{b}_L[j^*]$, $\mathbb{I}_{[0, L-1]} = \{0, 1, \ldots, L-1\}$, and $M$ represents a sufficiently large constant derived from the bounds on the pre-activation values. 
The first four constraints encode the ReLU activations using binary variables $\bm{z}_i$, while \cref{subeq:pixel_bounds} restricts the image pixels to their valid range. 
The remaining constraints define the group-wise patch attack. 
Since \cref{eq:gwise_verification_problem} introduces a binary variable for every ReLU unit, its computational cost grows rapidly with the network size, and directly applying this approach to large-scale neural network models is impractical.

\subsection{Incomplete Verification via SDP Relaxation}
\label{subsec:incomplete_sdp_relax}

We introduce our incomplete verification for the patch attack in this section. Firstly, we replace the ReLU representations, \cref{subeq:relu_constraint_2} and \cref{subeq:relu_constraint_end}, with quadratic constraints to enable SDP relaxation~\cite{Raghunathan2018sdp}.
By retaining the binary group-selection variable $\bm{G}$, the resulting mixed-integer semidefinite program (MISDP) preserves the exact group-wise patch attack model. Compared to \cref{eq:gwise_verification_problem}, this MISDP eliminates binary variables for individual neurons, drastically reducing the total number of binary decision variables from $\sum_{i=0}^{L-1} n_i$ to the number of groups $m$.

Following~\cite{Raghunathan2018sdp}, we introduce a vector 
\begin{align}
\tilde{\bm{x}} = (1, \bm{x}_0^\top, \bm{x}_1^\top, \ldots, \bm{x}_L^\top)^\top \in
\mathbb{R}^{1+\sum_{i=0}^{L} n_i}
\end{align}
 that concatenates $1$ and the activation vectors of all layers, and a symmetric matrix 
 \begin{align}
 P= \tilde{\bm x}\tilde{\bm x}^\top \in \mathcal{S}^{1+\sum_{i=0}^{L} n_i}.
 \end{align}
Then, \cref{eq:abstract_gwise_verification_problem} is equivalently reformulated with respect to the variable $P$ with a rank constraint $\operatorname{rank}(P) = 1$. 
By removing the rank constraint, we obtain the following MISDP:
\begin{subequations}
\label{eq:exact_misdp}
\begin{align}
\max_{P,\bm{G}}\quad &\bm{c}^\top P[\bm{x}_L]+c_0\\
  \text{s.t.}\quad &P[\bm{x}_{i+1}] \geq \bm 0 \quad (i \in \mathbb{I}_{[0,L-1]}),\\
  &P[\bm{x}_{i+1}] \geq W_i P[\bm{x}_i] + \bm{b}_i \quad (i \in \mathbb{I}_{[0,L-1]}),\\
  &\operatorname{diag}(
    P[\bm{x}_{i+1}\bm{x}_{i+1}^{\top}]
    - W_i P[\bm{x}_i \bm{x}_{i+1}^{\top}]) \nonumber\\
    &\quad - \bm{b}_i \odot P[\bm{x}_{i+1}] = \bm 0 \quad (i \in \mathbb{I}_{[0,L-1]}),\\
    &\operatorname{diag}\bigl(P[\bm{x}_i\bm{x}_i^\top]\bigr) \nonumber\\ 
  &\quad - (\bm{l}_i + \bm{u}_i) \odot P[\bm{x}_i] + \bm{l}_i \odot \bm{u}_i \leq \bm 0
  \quad (i \in \mathbb{I}_{[0,L]}), \\ 
    &\bm 0\le P[\bm{x}_0] \leq \bm 1, \label{subeq:misdp_pixel_bounds}\\
    &\| P[\bm{x}_0^{(g)}]- \bm{x}_0^{*(g)} \|_2 \leq \epsilon G^{(g)}\quad (g \in \mathcal{G}), \label{subeq:g_const_1}\\
    &\mathbf{1}^\top \bm{G} = k, \quad \bm{G} \in \{0,1\}^m, \label{subeq:g_const_2}\\
    &P[1] = 1,\quad P \succeq 0,
\end{align}
\end{subequations}
where we use the same indexing notation $P[\cdot]$ as in~\cite{Raghunathan2018sdp}, and $\bm{l}_i$ and $\bm{u}_i$ denote the lower and upper bounds of the pre-activation values of layer $i$, obtained by a lightweight method such as bound propagation~\cite{wang2021beta, zhang2018efficient}.
In addition, for the group-wise constraints we write $P[\bm{x}_0^{(g)}]$ for the sub-block of $P[\bm{x}_0]$ restricted to the index set $\mathcal{P}_g$ of group $g$.

\subsection{Proposed Relaxation for Group-Wise Sparsity}
\label{sec:relaxations}

The MISDP (\ref{eq:exact_misdp}) eliminates the binary variables associated with the ReLU activations, but retains the binary attack-selection variable $\bm{G}$. This remaining combinatorial structure still limits the scalability of the verification problem.
To address these limitations, we further relax the group-wise patch attack by removing the binary variable $\bm{G}$. Specifically, we construct a convex perturbation set $\widetilde{\mathcal{X}}$ satisfying
\begin{align}
\mathcal{X}\subseteq\widetilde{\mathcal{X}},
\label{eq}
\end{align}
where $\mathcal{X}$ is the original perturbation set defined in \cref{eq:gwise_perturb_set}. Replacing $\mathcal{X}$ with $\widetilde{\mathcal{X}}$ yields a relaxation of the original verification problem. Since the relaxed set contains the original attack set, a successful verification result remains sound for the original threat model.
We propose three relaxations with different levels of approximation: Full-Group, $L_2$-norm, and Group-Lasso relaxations. Specifically, we replace the constraints \cref{subeq:g_const_1} and \cref{subeq:g_const_2} with the following proposed constraints \cref{eq:full_group} - \cref{eq:group_lasso}.

\subsubsection{Full-Group Naive Relaxation}

The full-group relaxation simply removes the attack-selection constraint and allows a perturbation of radius $\epsilon$ in every group:
 \begin{align}
    \mathcal{X}_{\mathrm{Full}}
    = \left\{ \bm{x}_0 \;\middle|\;
        \| \bm{x}_0^{(g)} - \bm{x}_0^{*(g)} \|_2 \leq \epsilon,
        \quad \forall g \in \mathcal{G}
    \right\}. \label{eq:full_group}
\end{align}

This relaxation corresponds to fixing $G^{(g)}=1$ for all groups. Consequently, this completely removes the sparsity constraint, yielding the simplest convex approximation for the original attack set.

\subsubsection{$L_2$-Norm Relaxation}
The $L_2$-norm relaxation replaces the group-wise perturbation constraints with a single constraint on the entire image:
\begin{align}
    \mathcal{X}_{\mathrm{L_2}}
    = \left\{ \bm{x}_0 \;\middle|\;
        \| \bm{x}_0 - \bm{x}_0^* \|_2 \leq k\epsilon
    \right\}.
\end{align}
Although this relaxation method ignores the group structure of the attacks, it offers the advantage of reducing the problem to the standard $L_2$-norm ball formulation; consequently, existing SDP (semidefinite programming)-based verification frameworks for $L_p$-norm ball attacks can be applied directly.

\subsubsection{Group-Lasso Relaxation}
Group-Lasso relaxation preserves the group structure of the original attack while removing the binary selection variables. Specifically, we replace the individual group-wise constraints with a single group-sparse constraint:
\begin{align}
    \mathcal{X}_{\mathrm{GL}}
    = \left\{ \bm{x}_0 \;\middle|\;
        \sum_{g \in \mathcal{G}} \| \bm{x}_0^{(g)} - \bm{x}_0^{*(g)} \|_2 \leq k \epsilon
    \right\}. \label{eq:group_lasso}
\end{align}
This constraint bounds the total group-wise perturbation by the maximum perturbation that can be introduced into $k$ groups, while allowing the optimization to determine how the perturbation is distributed across groups. Among the three relaxations, Group-Lasso most directly reflects the spatial sparsity of the original group-wise patch attack.

\subsubsection{Soundness and Inclusion Relations}
\label{subsec:relaxation_inclusion}

The relationships among the original attack set and the proposed convex relaxations are formalized by the following propositions.

\begin{proposition}[Soundness]
\label{cor:soundness}
All three relaxed perturbation sets contain the original attack set $\mathcal{X}$. Specifically, we have the following relationships:
\begin{align}
\mathcal{X} \subseteq \mathcal{X}_{\mathrm{Full}}, \quad
\mathcal{X} \subseteq \mathcal{X}_{\mathrm{GL}}, \quad
\mathcal{X} \subseteq \mathcal{X}_{L_2}.
\end{align}
\end{proposition}

\begin{proof}
Let $\bm{x}_0\in\mathcal{X}$ be an arbitrary input satisfying \cref{eq:gwise_perturb_set}. For each group $g$,
\begin{align*}
    \| \bm{x}_0^{(g)}-\bm{x}_0^{\ast(g)} \|_2 \leq \epsilon G^{(g)} \leq \epsilon,
\end{align*}
which immediately gives $\bm{x}_0\in\mathcal{X}_{\mathrm{Full}}$.
Furthermore, 
\begin{align*}
\sum_{g} \| \bm{x}_0^{(g)} - \bm{x}_0^{*(g)} \|_2
= \sum_{g: G^{(g)} = 1} \| \bm{x}_0^{(g)} - \bm{x}_0^{*(g)} \|_2
\leq  k\epsilon,
\end{align*}
where the last inequality follows from $\mathbf{1}^{\top}\bm{G}=k$. Therefore, $\bm{x}_0\in\mathcal{X}_{\mathrm{GL}}$.
Finally, because the groups form a non-overlapping partition of the image,
\begin{align}
     \| \bm{x}_0 - \bm{x}_0^*\|_2
    \leq \sum_g \| \bm{x}_0^{(g)} - \bm{x}_0^{*(g)} \|_2
    \leq k\epsilon.
\end{align}
Hence, $\bm{x}_0\in\mathcal{X}_{L_2}$.
\end{proof}

\begin{proposition}[Inclusion Relations]
\label{cor:inclusion}
Group-Lasso relaxation is contained within the $L_2$-norm relaxation ($\mathcal{X}_{\mathrm{GL}} \subseteq \mathcal{X}_{L_2}$). Combined with the soundness properties, the resulting relationships can be summarized as:
\begin{align}
\mathcal{X} \subseteq \mathcal{X}_{\mathrm{GL}} \subseteq \mathcal{X}_{L_2}, \quad
\mathcal{X} \subseteq \mathcal{X}_{\mathrm{Full}}.
\label{eq:all_inclusions}
\end{align}
\end{proposition}

\begin{proof}
We can establish the inclusion relation between Group-Lasso and $L_2$-norm relaxations by using the triangle inequality. For any $\bm{x}_0\in\mathcal{X}_{\mathrm{GL}}$, we have:
\begin{align}
\| \bm{x}_0-\bm{x}_0^\ast \|_2
\leq \sum_{g\in\mathcal{G}} \| \bm{x}_0^{(g)}-\bm{x}_0^{\ast(g)} \|_2
\leq k\epsilon.
\end{align}
Thus, $\mathcal{X}_{\mathrm{GL}} \subseteq \mathcal{X}_{L_2}$. 
In contrast, no general inclusion relation exists between $\mathcal{X}_{\mathrm{Full}}$ and the other two relaxed sets, and their relative tightness depends on the number of groups $m$ and the number of attacked groups $k$.
\end{proof}

Thus, Group-Lasso provides a tighter relaxation than the $L_2$-norm relaxation while explicitly retaining the group structure of the original attack.
After replacing the original perturbation set $\mathcal{X}$ with any of the three convex sets above, the binary variable $\bm{G}$ is no longer required. The resulting verification problem is therefore a pure SDP with continuous variables only. This removes the combinatorial optimization over attack locations and enables substantially more scalable verification of larger neural networks.

\section{Numerical Experiments}
We evaluate the proposed relaxation methods from three perspectives: computational efficiency, relaxation tightness, and verification performance.

\subsection{Common Setup}

All experiments were conducted on a node with two AMD EPYC 9654 CPUs (96 cores per socket, up to 3707 MHz) and 768 GiB of memory. To solve the MISDP, we used SCIP-SDP~\cite{SCIP90} with SoPlex~\cite{gleixner2020linear} to handle the branch-and-bound framework required for solving the auxiliary LPs. For our methods, we used MOSEK~\cite{mosek} to solve the relaxed SDPs.
The image is partitioned into non-overlapping square groups using a sliding window, where the side length of each square group specifies the group size. When the side length of the image is not evenly divisible by the group size, the remaining pixels are grouped together as a single group.

\subsection{Evaluation of Computational Efficiency}
\label{subsec:scalability}

We evaluate the computational efficiency of the original MISDP and the three proposed relaxations. We measure both wall-clock computation time and memory consumption as the size of the underlying semidefinite matrix increases. This experiment is intended to quantify the computational benefit of eliminating the binary variables associated with the attack location.

\paragraph{Experimental Setup.}
We evaluate four network architectures, denoted \emph{Tiny}, \emph{Small}, \emph{Medium}, and \emph{Large}, as in \cref{tab:model_settings}. We vary the width of the hidden layers, which is the primary factor governing the dimension of the semidefinite matrix. Because the original $28 \times 28$ MNIST resolution is computationally prohibitive for complete verification baselines, we train and evaluate our models on downsampled $5 \times 5$ images ($25$ dimensions) of the MNIST training data. For the verification setup, we set the perturbation radius to $\epsilon = 0.1$, the target class to $5$, and attack cardinality to $k = 1$, partitioning the input into $4$-pixel groups.

\begin{table}[t]
\centering
\small
\caption{Model settings used for the scalability evaluation}
\label{tab:model_settings}
\begin{tabular}{lccc}
\toprule
     & Hidden layer  & $P$ matrix size & Accuracy \\
\midrule
Tiny   & 5 nodes $\times$ 2 layers  & $46\times46$ & 44.37\% \\
Small  & 10 nodes $\times$ 2 layers & $56\times56$ & 51.81\% \\
Medium & 15 nodes $\times$ 2 layers & $66\times66$ & 52.74\% \\
Large  & 20 nodes $\times$ 2 layers & $76\times76$ & 53.34\% \\
\bottomrule
\end{tabular}
\end{table}

\paragraph{Results.}

The computation time results are reported in \cref{tab:scalability}. Across all four evaluated network architectures, the proposed relaxations drastically accelerate verification compared to the exact MISDP formulation (\cref{eq:exact_misdp}), requiring at most 3\% of the computation time of the original MISDP. Specifically, the proposed relaxation methods achieve speedup factors of at least 41$\times$, 209$\times$, 1320$\times$, and 1430$\times$ for the \emph{Tiny}, \emph{Small}, \emph{Medium}, and \emph{Large} networks, respectively. This significant reduction in runtime stems from removing the binary group-selection vector $\bm{G}$.

\begin{table}[t]
\centering
\small
\caption{
Execution time comparison between MISDP (\ref{eq:exact_misdp}) and three proposed relaxations, across different network sizes. All proposed relaxations achieve <3\% of MISDP runtime.}
\label{tab:scalability}
\small
\begin{tabular}{l c c c c}
\toprule
\cmidrule(l){2-5}
Method & Tiny [s] & Small [s] & Medium [s] & Large [s] \\
\midrule
MISDP (\ref{eq:exact_misdp}) & 6.64  & 22.95 & 77.88 & 115.84 \\
\hdashline
Full-Group        & 0.161 & 0.11  & 0.057 & 0.080 \\
$L_2$ norm  & \textbf{0.032} & \textbf{0.06} & \textbf{0.055} & \textbf{0.074} \\
Group-Lasso & 0.036 & \textbf{0.06} & 0.059 & 0.081 \\
\bottomrule
\end{tabular}
\vspace{-0.1in}
\end{table}

\subsection{Evaluation of Relaxation Tightness}
\label{subsec:accuracy}

We investigate how closely the proposed relaxations approximate the original MISDP. We compare the objective values obtained by the three relaxations with those of the original MISDP. Because each relaxation enlarges the feasible perturbation set, its optimal objective value provides a measure of the relaxation tightness. 

\label{subsubsec:obj_increase}
\paragraph{Experimental Setup.}

We use the \emph{Small} network configuration from \cref{tab:model_settings}. The input consists of MNIST images compressed to 25 dimensions. We set the target class to 5, use the ground-truth class as the true class, set $k=1$, and use groups containing four elements.
We evaluate two perturbation radii, $\epsilon=0.1$ and $\epsilon=10^{-4}$. For each setting, we compare the objective value of each relaxation with that of the original MISDP. The reported increase is computed relative to the original MISDP objective value.

\begin{figure} [t]
    \centering
    \begin{tikzpicture}
    \begin{axis}[
        width=\linewidth,
        height=\linewidth,
        axis equal image,
        xlabel={Objective function value of original MISDP},
        ylabel={Objective function value of relaxation method},
        xmin=-0.6949, xmax=4.9723,
        ymin=-0.6949, ymax=4.9723,
        tick align=outside,
        tick style={black},
        axis lines=left,
        legend pos=north west,
        legend cell align={left},
        legend style={draw=black, fill=white, font=\small},
        every axis plot/.append style={line width=1pt},
    ]

    \addplot[dashed, gray, no markers, domain=-0.6949:4.9723, samples=2, forget plot]
        {x};

    \addplot[only marks, mark=o, mark options={scale=1.0}, color=blue]
        coordinates {
        (1.5518,2.0687) (1.5868,3.0779) (1.3139,1.9617) (2.3520,3.6265) (2.0737,3.3529) (1.3098,2.8943) (3.1740,4.6318) (0.0000,0.0000) (2.1383,3.7491) (1.3882,2.3192) (0.3264,1.8781) (3.5717,4.5631) (1.8038,2.6485) (0.8264,2.3307) (1.1684,2.7605) (0.6668,2.2175) (0.3754,1.9078) (0.0000,0.0000) (3.0765,4.0946) (1.2051,1.8645) (1.3401,2.4466) (0.8810,1.5620) (1.3826,2.3919) (0.3627,1.1552) (3.3525,4.4999) (3.1349,4.1034) (2.9439,3.9654) (1.1779,1.7438) (2.6648,3.8103) (1.3765,2.7871) (2.9196,3.9520) (2.0844,3.3364) (1.0737,1.8116) (2.8388,3.6944) (0.8953,1.9447) (0.8365,2.0373) (2.8802,3.8768) (-0.4949,1.0068) (3.0056,3.8932) (2.6082,3.6084) (0.3741,1.0559) (1.8650,3.1511) (3.1886,4.7531) (2.5584,3.5991) (3.1556,4.1558) (0.8409,1.9616) (1.9494,3.0268) (0.6376,1.8392) (2.7378,4.0470) (3.4275,4.7723)
        };
    \addlegendentry{Full-Group}

    \addplot[only marks, mark=triangle, mark options={scale=1.1}, color=red]
        coordinates {
        (1.5518,1.6675) (1.5868,1.8316) (1.3139,1.5131) (2.3520,2.7192) (2.0737,2.4939) (1.3098,1.5694) (3.1740,3.4114) (0.0000,0.0000) (2.1383,2.4364) (1.3882,1.7067) (0.3264,0.6145) (3.5717,3.8397) (1.8038,2.0473) (0.8264,1.0549) (1.1684,1.4388) (0.6668,0.9446) (0.3754,0.6566) (0.0000,0.0000) (3.0765,3.3468) (1.2051,1.4055) (1.3401,1.6380) (0.8810,0.9976) (1.3826,1.6952) (0.3627,0.5116) (3.3525,3.5595) (3.1349,3.4400) (2.9439,3.2124) (1.1779,1.3460) (2.6648,3.0514) (1.3765,1.5849) (2.9196,3.2786) (2.0844,2.3338) (1.0737,1.2966) (2.8388,3.0791) (0.8953,1.1650) (0.8365,1.1348) (2.8802,3.2061) (-0.4949,-0.2272) (3.0056,3.2577) (2.6082,2.8951) (0.3741,0.5809) (1.8650,2.2976) (3.1886,3.4809) (2.5584,2.8990) (3.1556,3.4228) (0.8409,1.1467) (1.9494,2.2534) (0.6376,0.9043) (2.7378,3.1403) (3.4275,3.6926)
        };
    \addlegendentry{$L_2$-Norm}

    \addplot[only marks, mark=square, mark options={scale=0.9}, color=green!55!black]
        coordinates {
        (1.5518,1.5518) (1.5868,1.5868) (1.3139,1.3139) (2.3520,2.3520) (2.0737,2.1380) (1.3098,1.3098) (3.1740,3.1740) (0.0000,0.0000) (2.1383,2.1383) (1.3882,1.4474) (0.3264,0.3264) (3.5717,3.5717) (1.8038,1.8084) (0.8264,0.8264) (1.1684,1.1684) (0.6668,0.6668) (0.3754,0.3754) (0.0000,0.0000) (3.0765,3.0765) (1.2051,1.2051) (1.3401,1.3401) (0.8810,0.8810) (1.3826,1.3827) (0.3627,0.3627) (3.3525,3.3525) (3.1349,3.1565) (2.9439,2.9439) (1.1779,1.1779) (2.6648,2.6814) (1.3765,1.3765) (2.9196,2.9289) (2.0844,2.0844) (1.0737,1.0929) (2.8388,2.8396) (0.8953,0.8953) (0.8365,0.8365) (2.8802,2.9041) (-0.4949,-0.4949) (3.0056,3.0062) (2.6082,2.6082) (0.3741,0.3746) (1.8650,1.9281) (3.1886,3.1886) (2.5584,2.5651) (3.1556,3.1556) (0.8409,0.8409) (1.9494,1.9494) (0.6376,0.6376) (2.7378,2.7688) (3.4275,3.4275)
        };
    \addlegendentry{Group-Lasso}

    \end{axis}
    \end{tikzpicture}
    \caption{Comparison of the objective values obtained by the original MISDP and the proposed relaxations for $\epsilon=0.1$. Each point corresponds to one evaluation sample. Group-Lasso is almost on the diagonal, demonstrating its very tight relaxation of the MISDP.}
    \label{fig:mokuteki_kansuu_hikaku}
\end{figure}
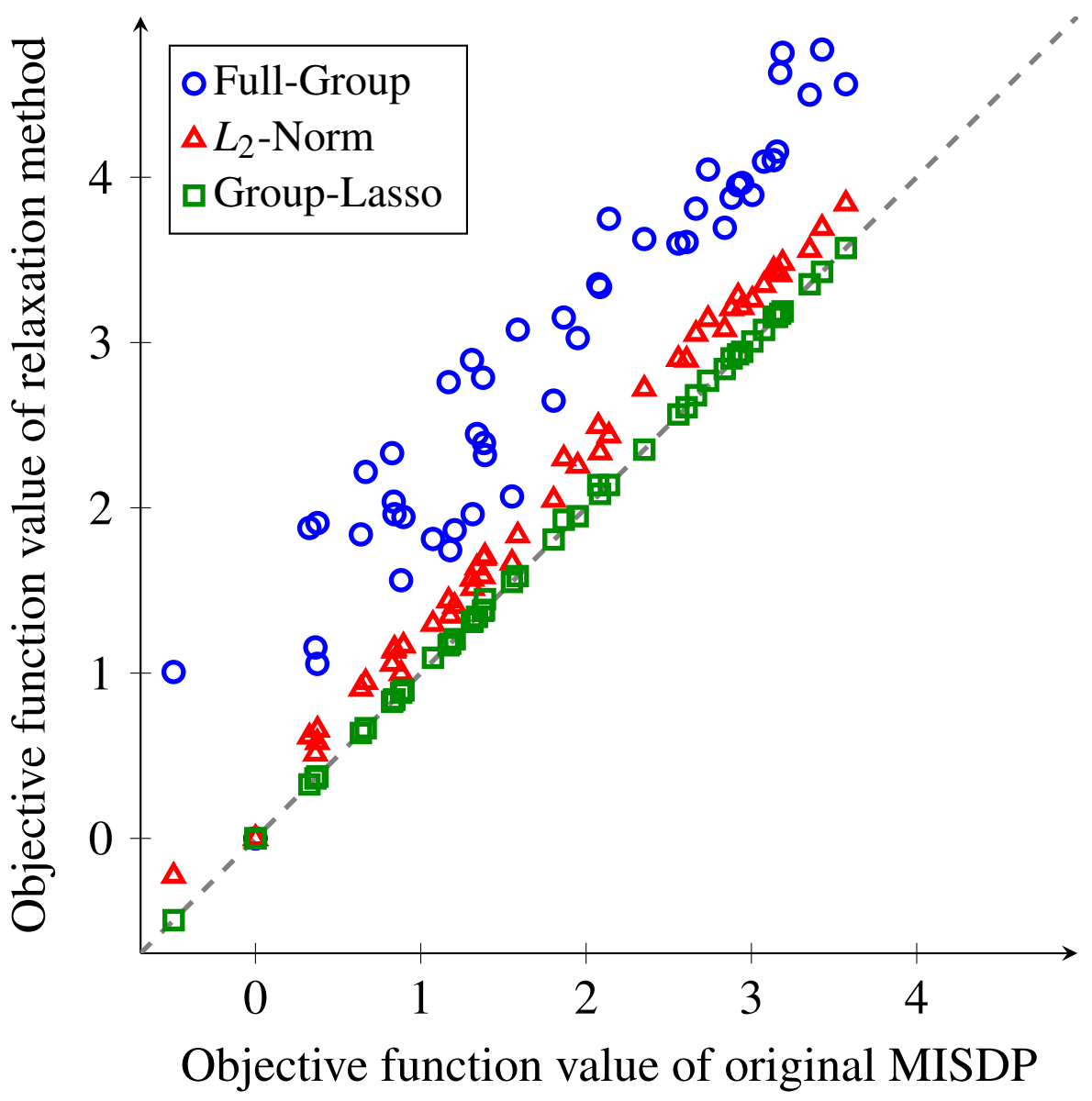

\paragraph{Results.}
\cref{tab:obj_increase} shows the average increase in the objective value relative to the original MISDP for different perturbation radii. 
For each perturbation radius, we randomly sampled 50 images from the MNIST test set, computed the percentage increase in the objective value relative to the original MISDP solution for each sample, and averaged the results.
As shown, Group-Lasso relaxation provides the tightest approximation to the original MISDP among the three relaxations.  Even at the large attack radius $\epsilon=0.1$, the objective value changes by only 0.34\% relative to the exact solution. This tight approximation is enabled by Group-Lasso relaxation, which can retain the group-wise structure of the original attack, whereas the other relaxations discard an increasing amount of this structure.
This trend is also evident in \cref{fig:mokuteki_kansuu_hikaku}. Group-Lasso solutions lie closest to the diagonal, while the $L_2$-norm and Full-Group relaxations exhibit progressively larger deviations. This ordering is consistent with the inclusion relations established in \cref{subsec:relaxation_inclusion}. 

\begin{table}[t]
\centering
\small
\caption{Average increase in the objective value relative to the original MISDP for different perturbation radii, averaged over 50 randomly sampled test images.}
\label{tab:obj_increase}
\begin{tabular}{lcc}
\toprule
Method & $\epsilon=0.1$ & $\epsilon=10^{-4}$ \\
\midrule
Full-Group & 92.82\% & 1.07\% \\
$L_2$-norm & 20.76\% & 0.19\% \\
Group-Lasso & 0.34\% & 0.00\% \\
\bottomrule
\end{tabular}
\end{table}

\subsection{Verification Rate Comparison}
\label{subsubsec:validation_success}

We evaluate the verification rates of the proposed relaxations, comparing them with prior verification methods in addition to the relaxation tightness comparison in \cref{subsubsec:obj_increase}.

\paragraph{Experimental Setup.}
We follow the evaluation methodology adopted in prior work~\cite{li2023sok} to facilitate comparison with existing verification methods. We evaluate two fully connected architectures, FCNNa and FCNNb. FCNNa consists of two hidden layers with width 20, while FCNNb consists of three hidden layers with width 100.
For both FCNNa and FCNNb, the input layer dimension matches the total number of pixels in the input image, which is 784 for MNIST in our case.
For each architecture, we consider both normally trained (\emph{Clean}) and PGD-adversarially trained (\emph{Adv.}) models. The adversarially trained models are trained using PGD with the perturbation radius specified by the corresponding setting.
For each setting, we evaluate 100 test samples, selected at indices $0,100,$..$,9900$.
The verification rate is the fraction of samples for which the verification optimization certifies that the model remains robust against the specified target class under the corresponding perturbation set.
We compare the proposed Full-Group, $L_2$-norm, and Group-Lasso relaxations against bounded-MILP~\cite{tjeng2019evaluating}, CROWN-IBP~\cite{zhang2020towards}, and CROWN~\cite{zhang2018efficient}. The attack cardinality $k$ is 1, partitioning the input into $4$-pixel groups.
These baselines are representative $L_p$-norm-based verification methods and include the state-of-the-art methods reported in \cite{li2023sok}.

\paragraph{Results.}
The verification rates across models and defense settings are summarized in \cref{tab:verification_results_mnist}. For the \emph{FCNNa} architecture, our proposed Group-Lasso relaxation achieves high verification rates of 86\% on the standard (Clean) model and 91\% on the adversarially trained (Adv.) model. Under the Adv. setting, Group-Lasso substantially outperforms existing baselines, including bounded-MILP (70\%), CROWN (47\%), and CROWN-IBP (2\%). Similarly, for the deeper \emph{FCNNb} network, Group-Lasso achieves the highest verification rate of 69\% in the Adv. setting, whereas prior baselines fail almost completely (bounded-MILP at 0\%, CROWN-IBP at 0\%, and CROWN at 4\%). The observation is consistent when $k$ is increased to 10, as shown in \cref{tab:verification_results_mnist_k10}. Although bounded-MILP and CROWN yield higher certified rates on the Clean \emph{FCNNb} model (84\%), we consider that our relaxations are particularly effective on challenging samples that are difficult to verify. 

To quantify this, we examine the margin of the optimal objective value $f(\bm{x})[j] - f(\bm{x})[j^*]$, where negative values closer to zero indicate narrower margins. Among the evaluated inputs, the average optimal value for samples certified \emph{exclusively} by Group-Lasso was $-1.69$, compared to $-10.94$ for samples certified by all methods. This significant gap indicates that conventional verifiers suffer from loose relaxation bounds on challenging inputs with small margins, whereas preserving the group-sparse attack structure allows Group-Lasso to certify such instances that remain unprovable by existing baselines.

\begin{table*}[t]
\centering
\small
\caption{Verification results on standard (Clean) and adversarially trained (Adv.) models. We compare our proposed SDP relaxations against prior verification baselines, showing state-of-the-art performance~\cite{li2023sok}.
}
\label{tab:verification_results_mnist}
\begin{tabular}{llcccc}
\toprule
\multirow{2}{*}{\textbf{Category}} & 
\multirow{2}{*}{\textbf{Method}} & 
\multicolumn{2}{c}{\textbf{FCNNa}} & 
\multicolumn{2}{c}{\textbf{FCNNb}} \\
\cmidrule(lr){3-4} \cmidrule(lr){5-6}
& & Clean ($\epsilon=0.02$) & Adv. ($\epsilon=0.10$)
& Clean ($\epsilon=0.02$) & Adv. ($\epsilon=0.30$) \\
\midrule
\multirow{4}{*}{Ours}
& Full-Group  & 83\% & 70\% & 59\% & 21\% \\
& $L_2$-norm  & \textbf{86\%} & 70\% & 59\% & \textbf{69\%} \\
& Group-Lasso & \textbf{86\%} & \textbf{91\%} & 69\% & \textbf{69\%} \\
\hline
\multirow{4}{*}{Baselines}
& Bounded-MILP & 80\% & 70\% & \textbf{84\%} & 0\% \\
& CROWN-IBP   & 77\% & 2\%  & 0\%  & 0\% \\
& CROWN       & 79\% & 47\% & \textbf{84\%} & 4\% \\
\bottomrule
\end{tabular}
\vspace{-0.1in}
\end{table*}

\begin{table}[t]
\centering
\small
\caption{Verification performance with $k=10$. Group-Lasso shows higher verification rates than other unstructured relaxations.}
\label{tab:verification_results_mnist_k10}
\begin{tabular}{lcc}
\toprule
Method & FCNNa Clean & FCNNa Adv. \\
\midrule
Full-Group       & 83\% & 70\% \\
$L_2$-norm          & 83\% & 66\% \\
Group-Lasso & 83\% & 80\% \\
\bottomrule
\end{tabular}
\vspace{-0.1in}
\end{table}

\subsection{Qualitative Analysis}
\label{subsec:qualitative}
\begin{figure}[t]
    \centering
    \begin{subfigure}[t]{0.59\linewidth}
        \centering
        \includegraphics[height=\subfigImgHeight]{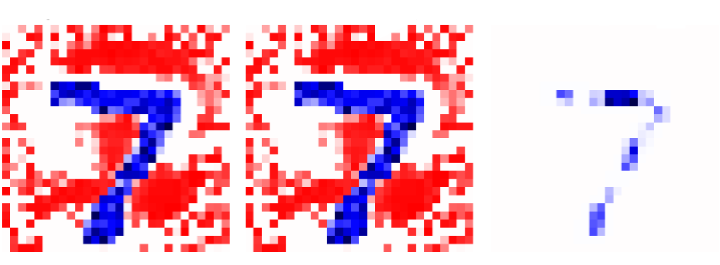}
        \caption{Ours: Full-Group, $L_2$, \& Group-Lasso}
        \label{fig:seven_delta_teian}
    \end{subfigure}
    \hfill
    \begin{subfigure}[t]{0.19\linewidth}
        \centering
        \includegraphics[trim={0cm 0cm 5cm 0cm},clip,height=\subfigImgHeight]{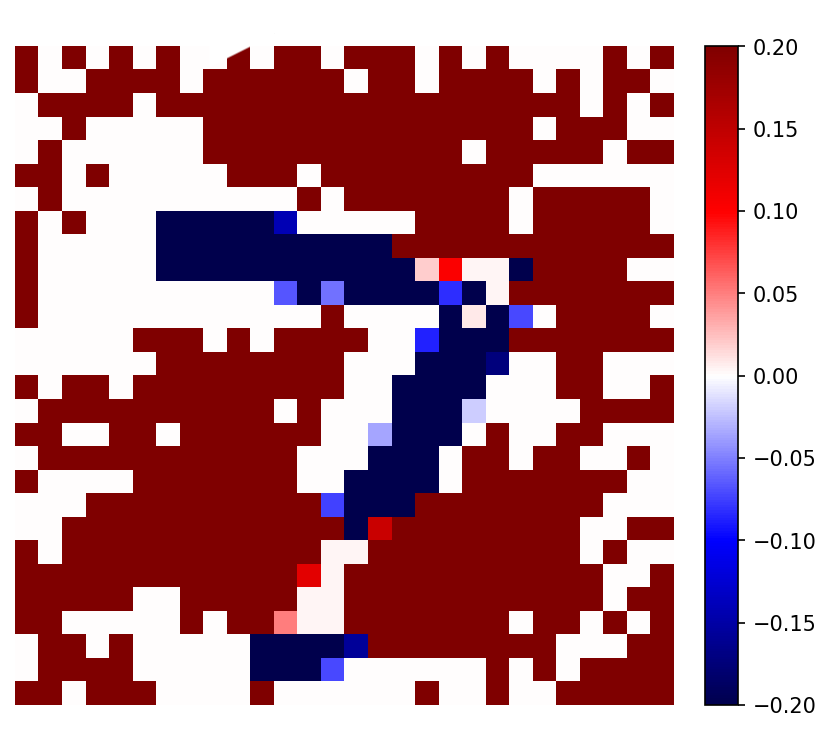}
        \caption{IBP}
        \label{fig:seven_delta_ibp}
    \end{subfigure}
    \hfill
    \begin{subfigure}[t]{0.19\linewidth}
        \centering
        \includegraphics[trim={0cm 0cm 5cm 0cm},clip,height=\subfigImgHeight]{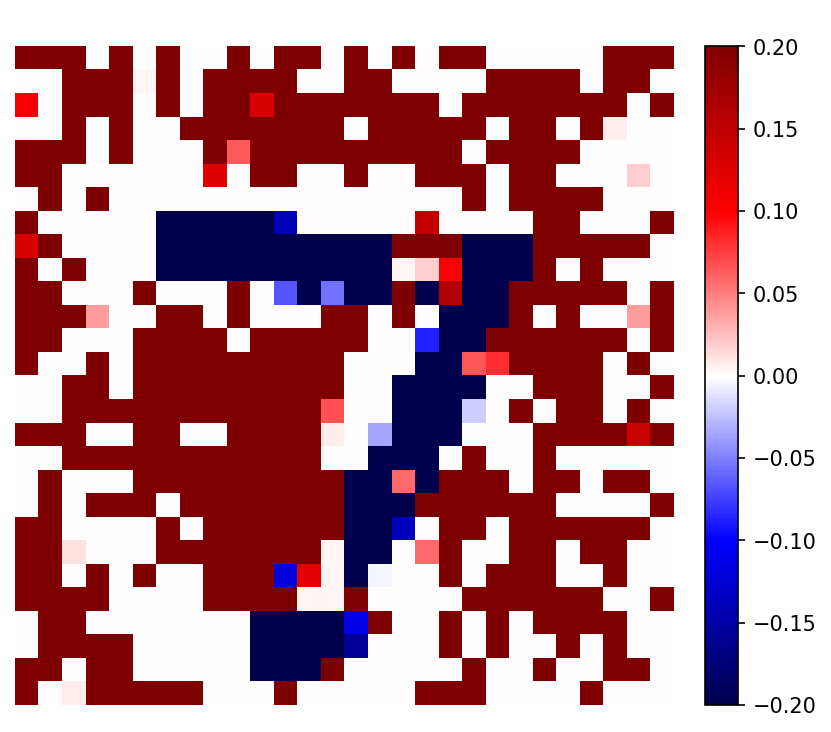}
        \caption{MILP}
        \label{fig:seven_delta_milp}
    \end{subfigure}
    \vspace{-0.05in}
    \caption{Visualization of perturbations of successfully verified samples when changing the predicted label from 7 to 0. Red and blue mean positive- and negative-direction perturbations, respectively. Only Group-Lasso can assume the localized perturbation.}
    \label{fig:seven_delta_all}
    \vspace{-0.1in}
\end{figure}

\begin{figure}[t]
    \centering
    \begin{subfigure}[t]{0.59\linewidth}
        \centering
        \includegraphics[height=\subfigImgHeight]{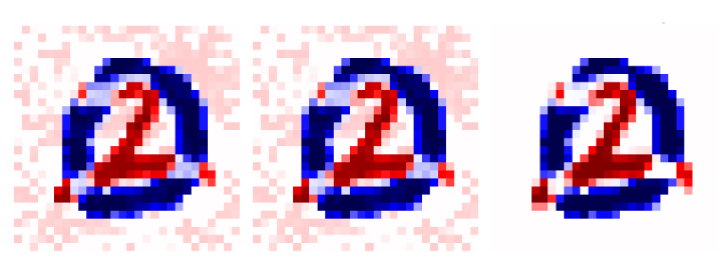}
        \caption{Ours: Full-Group, $L_2$, \& Group-Lasso}
        \label{fig:two_delta_teian}
    \end{subfigure}
    \hfill
    \begin{subfigure}[t]{0.19\linewidth}
        \centering
        \includegraphics[trim={0cm 0cm 5cm 0cm},clip,height=\subfigImgHeight]{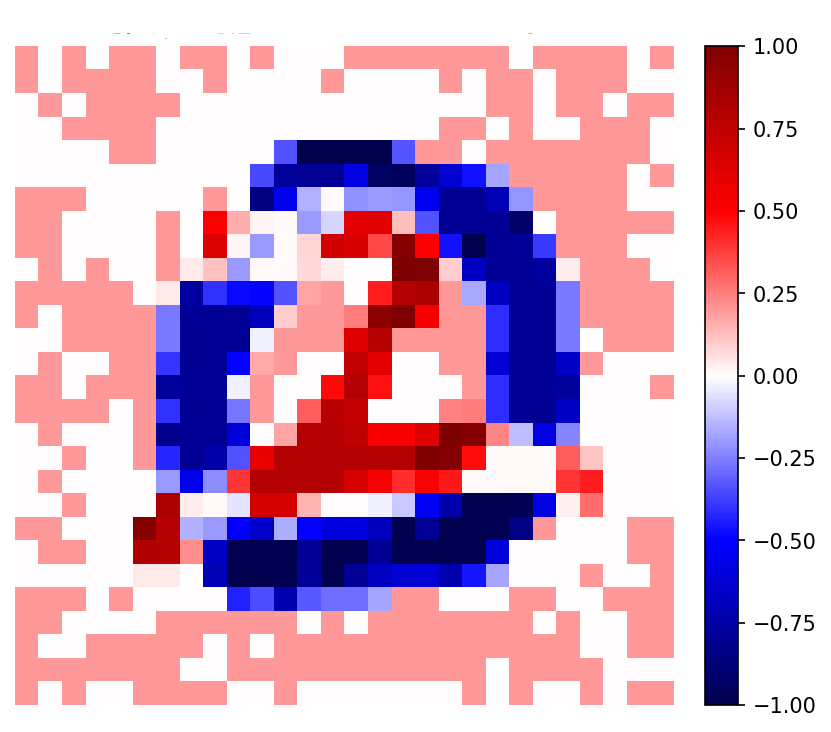}
        \caption{IBP}
        \label{fig:two_delta_ibp}
    \end{subfigure}
    \hfill
    \begin{subfigure}[t]{0.19\linewidth}
        \centering
        \includegraphics[trim={0cm 0cm 5cm 0cm},clip,height=\subfigImgHeight]{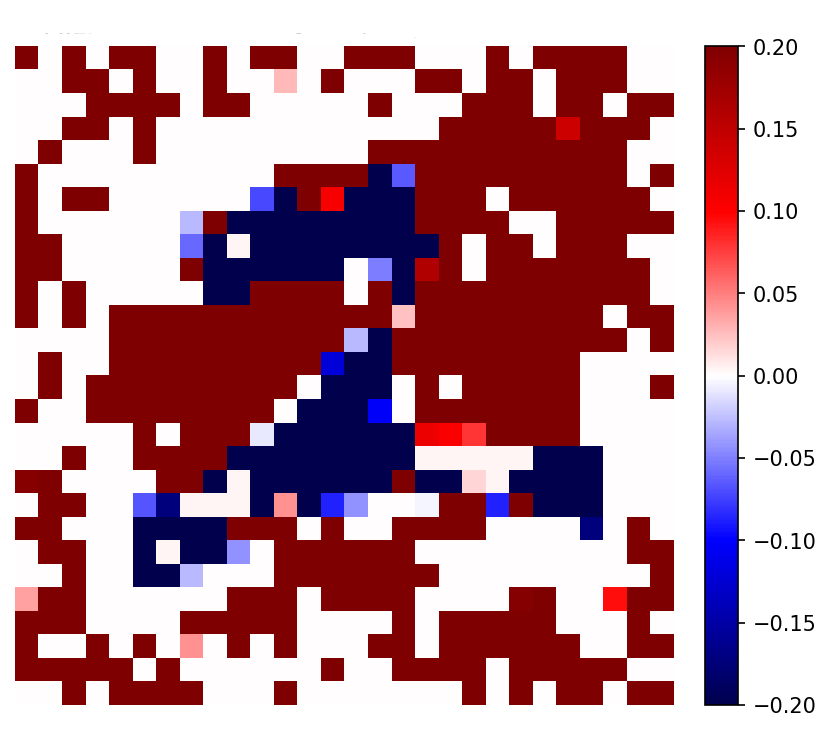}
        \caption{MILP}
        \label{fig:two_delta_milp}
    \end{subfigure}
    \vspace{-0.05in}
\caption{Visualization of perturbations for an unverified sample when changing the predicted label from 2 to 0. Red and blue mean positive- and negative-direction perturbations, respectively. Group-Lasso's perturbation is localized, but it attempts to draw 0.}
    \label{fig:two_delta_all}
    \vspace{-0.1in}
\end{figure}

\begin{figure}[t]
    \centering
    \setlength{\subfigImgHeight}{0.19\linewidth}%
    \newcommand{\panelGap}{\hspace{0.0573\subfigImgHeight}}%
    \begin{subfigure}[t]{3.097\subfigImgHeight}
        \centering
        \includegraphics[trim={10.5bp 9.75bp 2.25bp 18.75bp},clip,height=\subfigImgHeight]{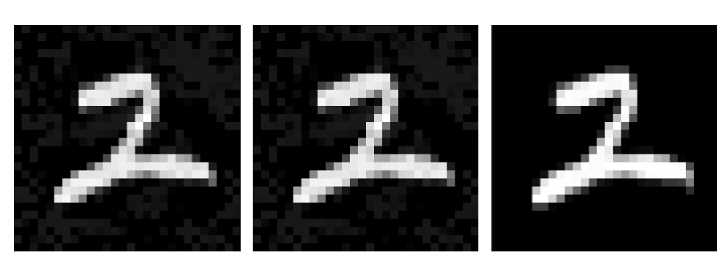}
        \caption{Ours: Full-Group, $L_2$, \& Group-Lasso}
        \label{fig:two_teian}
    \end{subfigure}%
    \panelGap%
    \begin{subfigure}[t]{\subfigImgHeight}
        \centering
        \includegraphics[trim={26bp 29bp 116bp 46bp},clip,width=\subfigImgHeight,height=\subfigImgHeight]{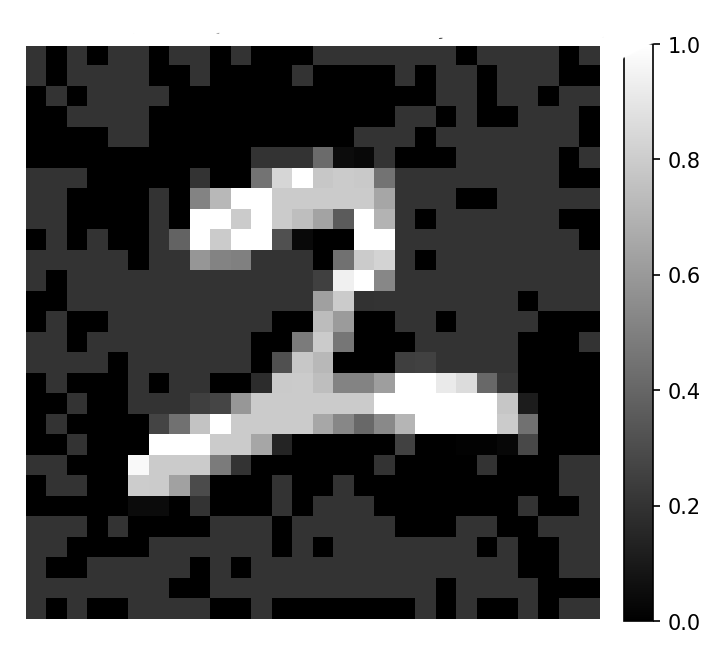}
        \caption{IBP}
        \label{fig:two_ibp}
    \end{subfigure}%
    \panelGap%
    \begin{subfigure}[t]{\subfigImgHeight}
        \centering
        \includegraphics[trim={16bp 29bp 117bp 46bp},clip,width=\subfigImgHeight,height=\subfigImgHeight]{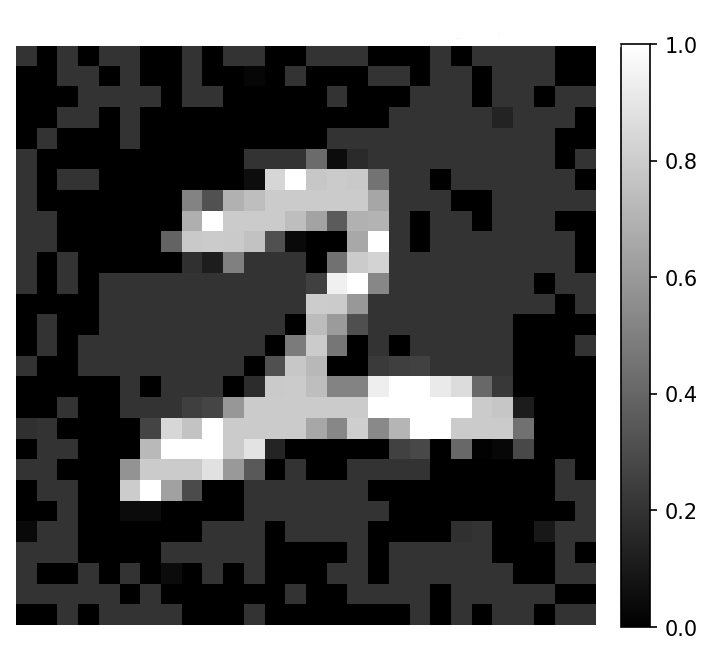}
        \caption{MILP}
        \label{fig:two_milp}
    \end{subfigure}
    \vspace{-0.05in}
\caption{Reconstructed attacked images from \cref{fig:two_delta_all}, attempting to change the prediction from 2 to 0.}
    \label{fig:two_all}
    \vspace{-0.1in}
\end{figure}

We qualitatively examine the perturbations obtained from the optimization solutions. This analysis provides an intuitive comparison between the perturbations produced by the proposed group-wise threat model and those produced by conventional $L_p$-based verification methods.

For bounded MILP, we directly use the optimal solution, as it is complete verification.
For incomplete verification, which only tests the upper bound of the objective value and does not return attack perturbations, we construct approximated attack perturbations using the following procedure.
For IBP perturbations, we determine the perturbation direction for each pixel based on its sensitivity to the true and target label scores, computed via forward propagation, and obtain an approximate perturbation by moving as far as possible in that direction within the input perturbation bound.
For SDP-based methods including ours, we extract the sub-vector corresponding to the original image and reshape it back into the original image's 2D array. Specifically, the sub-vector extracted corresponds to entries [0,1] through [0,256] of the SDP solution.

\vspace{-0.08in}
\paragraph{Perturbations of a Successfully Verified Sample.}
\Cref{fig:seven_delta_all} shows the perturbations obtained for an input image whose true label is 7, shown in \cref{fig:milp_ibp_glasso}, with target class 0, in a setting where Group-Lasso successfully verifies the model. All methods attempt to suppress or modify the digit 7. However, Group-Lasso relaxation concentrates the perturbation almost entirely on the digit itself, leaving the surrounding background unchanged. This behavior is consistent with the assumed localized patch threat model.
In contrast, the perturbations obtained by the other methods extend into the background region. In particular, the perturbations produced by IBP and MILP distribute non-negligible changes outside the digit region. The proposed Full-Group and $L_2$-norm relaxations reduce this effect compared with these baselines, but still permit perturbations outside the localized region. These observations illustrate the benefit of explicitly retaining group sparsity in the perturbation set.

\vspace{-0.08in}
\paragraph{Perturbations of an Unverified Sample.}
\Cref{fig:two_delta_all} shows an instance with true label 2, shown in \cref{fig:milp_ibp_glasso}, and target class 0 for which Group-Lasso does not certify robustness. \cref{fig:two_all} shows the corresponding reconstructed images. In this case, the perturbations produced by most methods modify both the digit and the surrounding background. Group-Lasso not only concentrates its perturbation on the digit region, but also tries to draw the target digit, 0. This supports the high capability of Group-Lasso to handle the patch attack, which can only inject localized perturbations. When subtle perturbations are insufficient, it is natural that the attacker instead draws the target digit.

In summary, the qualitative results support the motivation for the proposed relaxation: Group-Lasso not only provides a substantially tighter relaxation than the simpler alternatives, but also produces optimization solutions whose perturbations are spatially concentrated in a manner consistent with the assumed patch attack.

\section{Discussion and Limitations}
\label{sec:discussion}
The results highlight the importance of preserving the spatial structure of patch threats during verification. Among the proposed relaxations, Group-Lasso consistently provides the tightest approximation to the original MISDP and effectively concentrates perturbations within targeted regions, leading to stronger robustness certification against patch attacks.
However, there are several limitations in this study. We evaluated only fully connected networks and small-scale inputs, leaving high-resolution images and modern architectures for future work. Our primary focus is the theoretical formulation of patch attacks (\cref{sec:methodology}) and the tightness of the proposed relaxations relative to the original MISDP, which is itself unsuitable for large-scale problems. As the first formal verification framework specialized for patch attacks, we leave the development of scalable solvers, such as first-order methods~\cite{ahmadi2019dsos,burer2003nonlinear}, to future work. Finally, our fixed, non-overlapping image partition may not effectively capture arbitrary physical patch geometries. Nevertheless, it can flexibly incorporate realistic deployable regions, which are naturally constrained in physical settings (e.g., stickers cannot be placed on the sky).

\section{Conclusion}
\label{sec:conclusion}

In this work, we introduced a one-shot framework for robustness verification against patch attacks. Instead of enumerating possible attack locations, we formulated the attack location jointly with the adversarial perturbation and derived an exact mixed-integer semidefinite programming formulation. We then eliminated the discrete attack-selection variables through three convex relaxations: Full-Group, $L_2$-norm, and Group-Lasso. Our analyses demonstrate that explicitly preserving the spatial group-sparsity of real-world threats is critical for effective verification. Group-Lasso relaxation provides a tight and efficient approximation, incurring only a 0.34\% average increase in the objective value compared to the original MISDP while requiring less than 3\% of the computation time. Furthermore, by achieving a verification rate of 91\% compared to $\leq$70\% for prior baselines, our approach offers a practical, scalable solution for certifying neural network robustness. Future work will explore extending this framework to broader classes of physical attacks and larger network architectures.

{
    \small
    \bibliographystyle{plainnat}
    \bibliography{main}
}

\end{document}